\documentclass[10pt,twocolumn,letterpaper]{article}

\usepackage{iccv}
\usepackage{times}
\usepackage{epsfig}
\usepackage{graphicx}
\usepackage{amsmath}
\usepackage{amssymb}

\usepackage{multirow}
\usepackage[american]{babel}
\usepackage{array}
\usepackage{booktabs}
\usepackage[abs]{overpic}
\usepackage{nccmath}
\usepackage{bbding}
\usepackage[accsupp]{axessibility}

\usepackage[pagebackref=true,breaklinks=true,letterpaper=true,colorlinks,bookmarks=false]{hyperref}

\usepackage[capitalize]{cleveref}
\crefname{section}{Sec.}{Secs.}
\Crefname{section}{Section}{Sections}
\Crefname{table}{Table}{Tables}
\crefname{table}{Tab.}{Tabs.}
\Crefname{equation}{Equation}{Equations}
\crefname{equation}{Eqn.}{Eqns.}

\iccvfinalcopy 

\def\iccvPaperID{****} 

\ificcvfinal\fi

\begin{document}

\title{Beyond Image Borders: Learning Feature Extrapolation for\\Unbounded Image Composition}

\author{Xiaoyu Liu$^1$, Ming Liu$^{1(}$\Envelope$^)$, Junyi Li$^1$, Shuai Liu, Xiaotao Wang, Lei Lei, Wangmeng Zuo$^{1,2}$ \\
$^1$Harbin Institute of Technology, Harbin, China \quad $^2$ Peng Cheng Laboratory, Shenzhen, China\\
{\tt\small 
\href{mailto:liuxiaoyu1104@gmail.com}{\color{black}liuxiaoyu1104@gmail.com}, \href{mailto:csmliu@outlook.com}{\color{black}csmliu@outlook.com}, \href{mailto:nagejacob@gmail.com}{\color{black}nagejacob@gmail.com}, \href{mailto:wmzuo@hit.edu.cn}{\color{black}wmzuo@hit.edu.cn}}
}

\maketitle
\ificcvfinal\thispagestyle{empty}\fi

\begin{abstract}
For improving image composition and aesthetic quality, most existing methods modulate the captured images by striking out redundant content near the image borders.
However, such image cropping methods are limited in the range of image views.
Some methods have been suggested to extrapolate the images and predict cropping boxes from the extrapolated image.
Nonetheless, the synthesized extrapolated regions may be included in the cropped image, making the image composition result not real and potentially with degraded image quality. 
%
%
In this paper, we circumvent this issue by presenting a joint framework for both unbounded recommendation of camera view and image composition (i.e., UNIC).
In this way, the cropped image is a sub-image of the image acquired by the predicted camera view, and thus can be guaranteed to be real and consistent in image quality.  
Specifically, our framework takes the current camera preview frame as input and provides a recommendation for view adjustment, which contains operations unlimited by the image borders, such as zooming in or out and camera movement.
To improve the prediction accuracy of view adjustment prediction, we further extend the field of view by feature extrapolation.
After one or several times of view adjustments, our method converges and results in both a camera view and a bounding box showing the image composition recommendation.
Extensive experiments are conducted on the datasets constructed upon existing image cropping datasets, showing the effectiveness of our UNIC in unbounded recommendation of camera view and image composition.
The source code, dataset, and pre-trained models is available at \url{https://github.com/liuxiaoyu1104/UNIC}.

\end{abstract}

\vspace{-1.0em}
\section{Introduction}
\label{sec:BIB_introduction}
\vspace{-0.2em}

With the prevalence of electronic devices such as smartphones, taking photos has become a common activity in everyday life.
Due to the lack of professional photography knowledge and skills, taking photos with harmonious image composition and high aesthetic quality is still difficult for non-professional users.
As a remedy, image composition, which aims to find an aesthetic region of a scene, has attracted much attention in recent years.

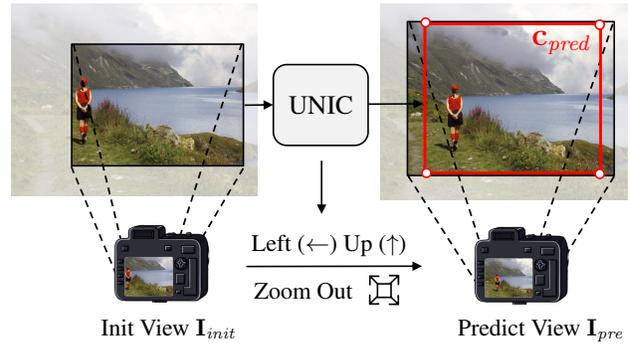
\begin{figure}[!t]
\centering
 \begin{overpic}[width=0.99\linewidth]{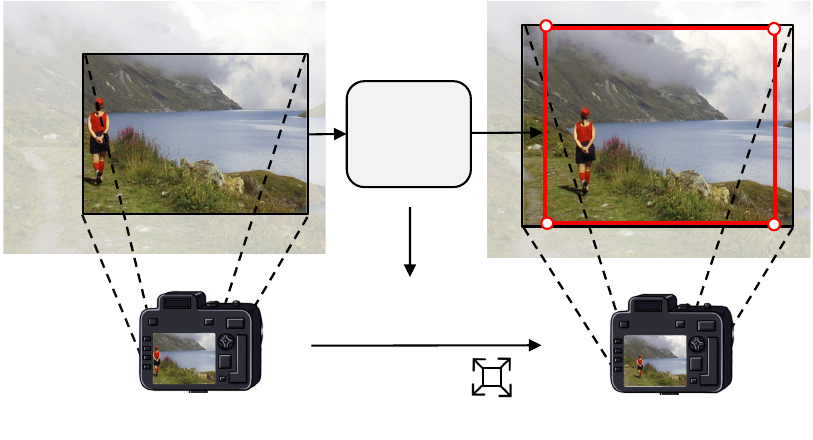}
  \put(106,85){\normalsize{UNIC}}
  \put(92,33){\small{Left ($\leftarrow$) Up ($\uparrow$)}}
  \put(93,15){\small{Zoom Out}}
  \put(35,1){\small{Init View $\mathbf{I}_\mathit{init}$}}
  \put(170,1){\small{Predict View $\mathbf{I}_\mathit{pre}$}}
  \put(198,111){\large\textcolor{red}{{$\mathbf{c}_\mathit{pred}$}}}

\end{overpic}
\vspace{1mm}
    \caption{Illustration of our proposed UNIC for unbounded recommendation of camera view and image composition. On the left is the initial view provided by the user. Given the current view, our model can predict camera operations (\eg, zoom out and the movement) and a image composition solution (\eg, $\mathbf{c}_\mathit{pred}$). The prediction can be executed multiple times until convergence.}
    \label{fig:BIB_introdution}
    \vspace{-1em}
\end{figure}

In order to facilitate the training of image composition models, several datasets~\cite{wei2018good,zeng2019reliable,chen2017quantitative} have been established, and each image comes along with one or multiple bounding boxes indicating the cropping schemes.
Despite the notable progress, most existing image composition methods~\cite{wei2018good,guo2018automatic,li2018a2,li2019fast,zeng2019reliable,lu2019end,zhong2021aesthetic,hong2021composing} generally adopt a post-processing form on the already captured images, \ie, they only adjust the composition in an image cropping manner.
In other words, the captured images are modulated by striking out redundant content near the image borders.
Nonetheless, a sub-optimal solution will inevitably be obtained when the best cropping is not entirely in the acquired image.
%
%
To alleviate this issue, Zhong~\etal~\cite{zhong2021aesthetic} proposed to expand the image via out-painting, and then predict the cropped view on the expanded image.
%
It is a practical solution in post-processing manner,
but may suffers from outpainting artifacts.
%

To tackle the limitations of existing image composition methods, this paper proposes a novel framework for \textbf{un}bounded recommendation of camera view and \textbf{i}mage \textbf{c}omposition (\ie, UNIC).
As shown in \cref{fig:BIB_introdution}, the user initializes a view with the content of interest.
Given the current view, our model finds a potential well-composed view and provides the corresponding camera movement operations, either inside or beyond the image borders.
Note that solely performing camera view adjustment is not enough, since the aspect ratio is typically kept unchanged during the photography process.
Therefore, our model also concurrently predicts a bounding box for cropping after camera movement operations.
With our model, the user finally can get the most recommended camera view and the corresponding image composition bounding box as shown in \cref{fig:BIB_introdution}.

For implementing the UNIC model, we further simplify the task of joint camera view adjustment and image composition into unbounded image composition by merging the outputs.
In this way, the architecture of cropping based image composition methods can be deployed as the backbone, and we follow Jia~\etal~\cite{jia2022rethinking} to adopt the conditional-DETR~\cite{meng2021conditional} structure.
In contrast to existing image cropping methods, we argue that our UNIC is more preferred and practical.
%
%
First, existing image composition methods~\cite{wei2018good,guo2018automatic,li2018a2,li2019fast,zeng2019reliable,lu2019end,hong2021composing} are restricted to image cropping over the already captured images.
The introduction of camera view adjustment can naturally circumvent the restriction of image borders by moving the camera or adjusting the optical zoom. 
Second, in comparison to out-painting~\cite{zhong2021aesthetic}, camera view adjustment can guarantee that the pixels outside the original borders are real and consistent with the pixels within the borders.
Furthermore, new and real information can be introduced after each time of view adjustment. 
Thus, based on the result of last time, one can perform view adjustment and image composition for many times, which is also not supported by image cropping based methods.

Moreover, our UNIC is free to go beyond image borders, yet directly predicting in unseen regions may lead to inferior results.
To compensate for this, we further extend the camera field of view by extrapolation.
Different from Zhong~\etal~\cite{zhong2021aesthetic}, whose extrapolation was performed in the image domain, we choose feature extrapolation and use it for predicting camera movement and bounding box  instead of synthesizing unseen content.
Thus, we can get the content in novel views by moving the camera, and unseen content generation is not necessary.
In comparison to the latent space specified for image composition, forcing the extrapolation into the image domain may bring redundant or even harmful information.
Besides, the feature extrapolation module can be well integrated into our existing framework, avoiding the heavy computation burdens brought by extra modules such as the image decoder.

For training and evaluating the proposed model, we take the advantage of existing image cropping datasets~\cite{wei2018good,zeng2019reliable} and convert them into a more generalized form.
Extensive experiments and ablation studies show the effectiveness of our UNIC, which can work well under diverse conditions.

To sum up, the contributions of this paper include,
\begin{itemize}
    \item  We propose a novel UNIC method for jointly performing unbounded recommendation of camera view and image composition. The user can adjust the current view following the recommendations to obtain images with higher aesthetic quality.
    \item We introduce a feature extrapolation module as well as an extrapolation loss term in the detection transformer framework, which improves the prediction accuracy, especially for out-of-image scenarios.
    \item Two unbounded image composition datasets are constructed upon existing image cropping ones. Experimental results show that our proposed method achieves superior performance against state-of-the-art methods.
\end{itemize}


\section{Related Work}
\label{sec:BIB_related_work}

\subsection{Image Composition}
%
Image composition aims to find the most aesthetic photo of a scene, which is typically achieved by image cropping in the literature.
Early works rely on saliency detection to preserve important content in the image~\cite{chang2009finding,fang2014automatic,chen2016automatic,sun2013scale} or extract hand-crafted features from aesthetic characteristics or composition rules for predicting cropping schemes~\cite{chang2009finding,cheng2010learning,zhang2012probabilistic,su2012preference,yan2013learning,zhang2013weakly}.
Recently, a large number of methods address image cropping tasks in a data-driven manner.
In general, existing methods can be broadly categorized as two groups, \ie, anchor evaluation~\cite{chen2017learning,wei2018good,zeng2019reliable,zhong2021aesthetic,li2020learning} and cropping coordinate regression~\cite{guo2018automatic,lu2019end,li2018a2,li2019fast,hong2021composing}.

\vspace{0.5em}
\noindent\textbf{Anchor Evaluation.}
The general pipeline of anchor evaluation based methods is to generate candidate croppings and then rank different crops to obtain the final result.
For example, Chen~\etal~\cite{chen2017learning} proposed paired ranking constraints to train an aesthetics-aware ranking net.
Wei~\etal~\cite{wei2018good} predicted scores efficiently by introducing a new knowledge transfer framework.
Zeng~\etal~\cite{zeng2019reliable} constructed a novel grid anchor based formulation and a corresponding dataset for image cropping.
CGS~\cite{li2020composing} explicitly utilized mutual relations between different candidate regions with a graph-based model.
Besides, two tasks closely related to our method are worth mentioning. Zhong~\etal~\cite{zhong2021aesthetic} expanded the range of cropping windows outside the image border through image out-painting. 
However, the out-painting result may suffer from low visual quality and be inconsistent with the real-world scene.
And some methods~\cite{rawat2015context,lou2021aesthetic} also tried to provide composition scores for the current view when photographing with mobile devices, yet they lack the ability to recommend new camera views.

\vspace{0.5em}
\noindent\textbf{Coordinate Regression.}
Coordinate regression based methods directly obtain the coordinate of the cropping box.
Some works~\cite{guo2018automatic,lu2019end} directly designed an end-to-end network to predict the cropping boxes.
Regarding image cropping as a consistent decision-making process, Li~\etal~\cite{li2018a2,li2019fast} introduced reinforcement learning to generate boxes.
Composition rules were explicitly leveraged by Hong~\etal~\cite{hong2021composing}, making the model work like a photographer.
Based on object detection method~\cite{meng2021conditional}, Jia~\etal~\cite{jia2022rethinking} predicted multiple crop schemes in a set prediction manner, which took model diversity and globalization into account.

In comparison to the aforementioned cropping based image composition methods, our solution performs unbounded recommendation of camera view and image composition, which can provide more flexibility in searching for better composition schemes.

\subsection{Image Out-painting}
In this work, we extrapolate the features for better prediction, which is closely related to image out-painting methods.
Therefore, we briefly review the progress of out-painting tasks.
Inspired by image in-painting methods, Sabini and Rusak~\cite{sabini2018painting} introduced the image out-painting task and trained a deep neural network framework adversarially.
Wang~\etal~\cite{wang2019wide} designed an effective deep generative model termed SRN with practical context normalization module for image extrapolation.
Some spatial-related loss terms are also proposed to improve the performance.
For example, a recurrent content transfer model was proposed for spatial content prediction in NSIPO~\cite{yang2019very}.
Based on StyleGAN2~\cite{karras2020analyzing}, Zhao~\etal~\cite{zhao2021large} presented comodulated GANs, which utilized the difference between the unconditional and conditional generative models.
Moreover, Ma~\etal~\cite{ma2021boosting} decomposed the image out-painting task into two generation stages, \ie, semantic segmentation domain and image domain.
More recently, transformer-based networks are incorporated to extend image borders.
For example, Gao~\etal~\cite{gao2022generalised} designed a transformer-based generative adversarial network with Swin transformer blocks~\cite{liu2021swin}.
QueryOTR~\cite{yao2022outpainting} proposed a novel hybrid transformer that formulated out-painting problem as a sequence-to-sequence auto-regression problem.
In this work, we extrapolate in the feature domain, which shows superior performance against image out-painting for our UNIC.


\begin{figure*}[htp]
    \centering
    \begin{overpic}[percent,width=.99\linewidth]{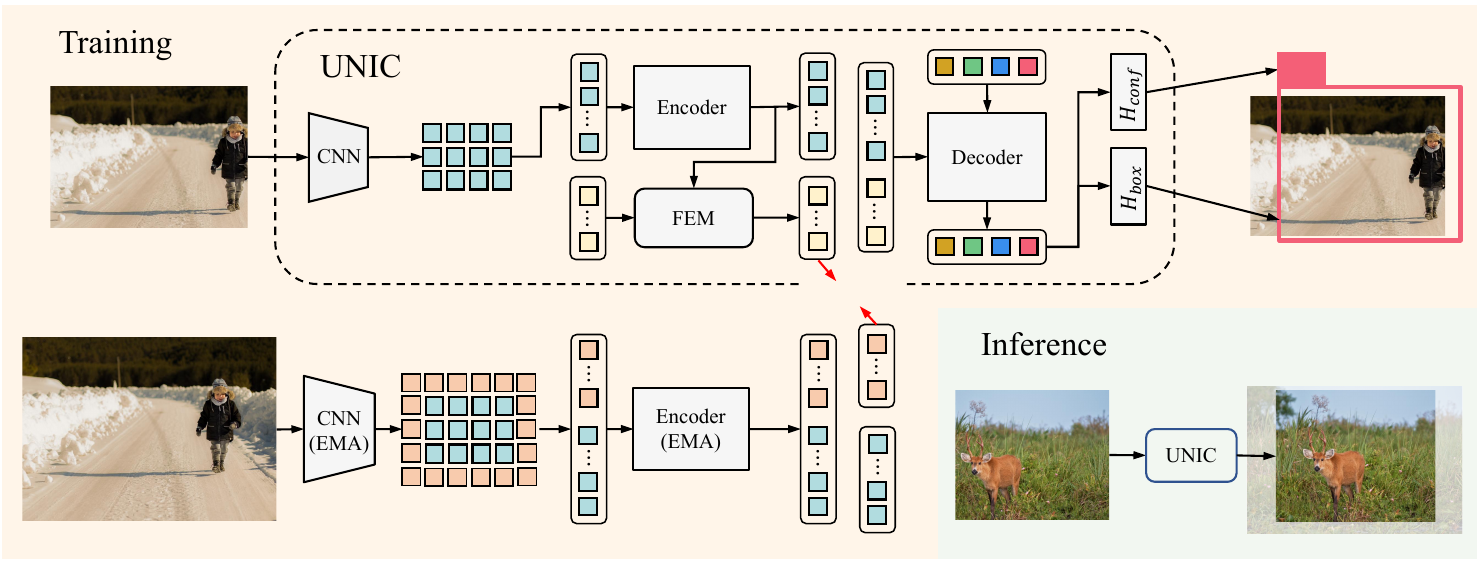}
    \put(55,17.7){\textcolor{red}{\small{$\mathcal{L}_{\text{extra}}$}}}
     \put(9.36,20.12){\small{$\mathbf{I}_{\text{init}}$}}
     \put(10,1.0){\small{$\mathbf{I}$}}
     \put(30.5,22.75){\small{$\mathbf{h}_{\text{init}}$}}
     \put(31.68,2.9){\small{$\mathbf{h}$}}
     \put(51,31.5){\tiny{$\mathcal{Z}_\mathit{vis}$}}
     \put(51,24){\tiny{$\mathcal{Z}_\mathit{pad}$}}
     \put(60.6,13){\tiny{$\mathcal{Z}_\mathit{out}$}}
     \put(60.75,4.5){\tiny{$\mathcal{Z}_\mathit{in}$}}
     \put(79.9,30.5){\small{$\mathbf{p}_\mathit{pred}$}}
     \put(79.8,26){\small{$\mathbf{c}_\mathit{pred}$}}
     \put(71.1,32.8){\small{$\mathcal{A}$}}
     \put(86.5,32.7){\textcolor{white}{\scriptsize{$0.89$}}}
    \end{overpic}
    \caption{Architecture of the proposed UNIC framework. It adopts a cDETR-like encoder-decoder architecture~\cite{meng2021conditional} to predict aesthetic plausible view $\mathbf{c}_\mathit{pred}$ from initial view $\mathbf{I}_\mathit{init}$. To mitigate the difficulty in predicting $\mathbf{c}_\mathit{pred}$ beyond image borders, a feature extrapolation module is deployed to predict the invisible tokens $\mathcal{Z}_\mathit{pad}$ from visible ones $\mathcal{Z}_\mathit{vis}$. The FEM is supervised by tokens extracted from larger view $\mathbf{I}$ with the exponential moving averaged CNN and encoder during training.}
    \label{fig:BIB_method}
    \vspace{-0.5em}
\end{figure*}
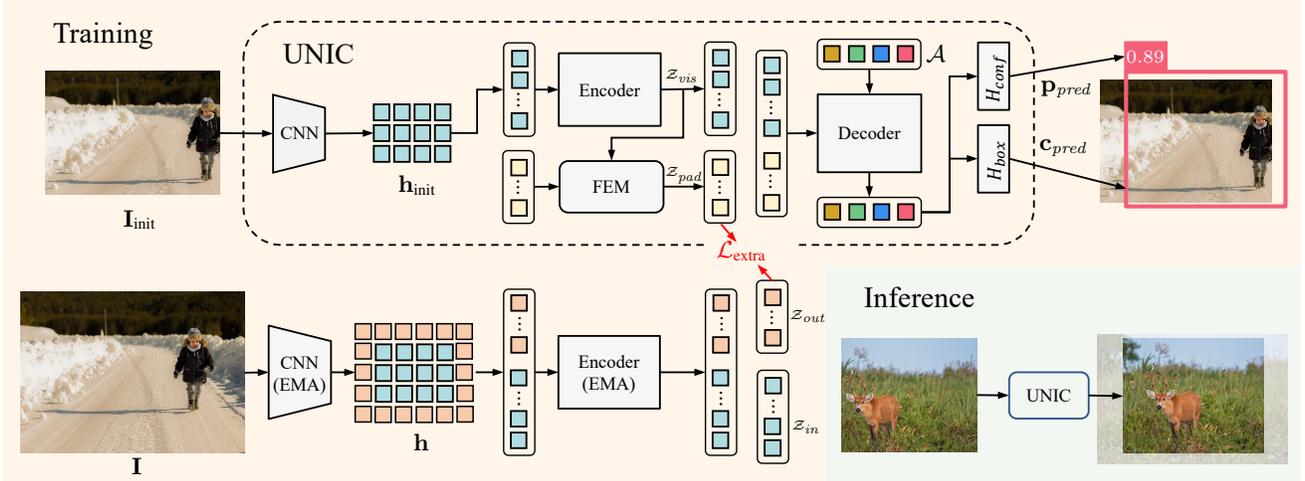
\section{Method}

\subsection{Problem Definition and Overview}

While existing cropping-based image composition methods predict a bounding box for image cropping, we extend the problem to joint unbounded recommendation of camera view and image composition.
In specific, the user initializes a camera view $\mathbf{I}_\mathit{init}$  with field of view $\mathbf{v}_\mathit{init}$\footnote{We represent the position and size of bounding boxes by four values $[\mathit{x}, \mathit{y}, \mathit{w}, \mathit{h}]$, where $(\mathit{x}, \mathit{y})$ is the center coordinate, while $\mathit{w}$ and $\mathit{h}$ are width and height, respectively. The axes are normalized to $[0, 1]$ \wrt $\mathbf{I}_\mathit{init}$.}, which contains the subjects or scenes of interest.
Given $\mathbf{I}_\mathit{init}$, we predict the actions (\eg, zoom in/out, move left/right, move up/down, \etc) for obtaining a new camera view $\mathbf{I}_\mathit{pred}$ located by $\mathbf{v}_\mathit{pred}$.
{In practice, the view with a high aesthetic score may not share the same aspect ratio as the camera, thus we concurrently predict a bounding box (denoted by $\mathbf{c}_\mathit{pred}$) for cropping in the adjusted camera view.}
With a model $\mathit{f}(\cdot)$, the problem can be formulated by
\begin{equation}
    [\mathbf{v}_\mathit{pred}, \mathbf{c}_\mathit{pred}] = \mathit{f}(\mathbf{I}_\mathit{init}),
    \label{eqn:BIB_definition}
\end{equation}
where the difference between $\mathbf{v}_\mathit{pred}$ and $\mathbf{v}_\mathit{init}$ indicates the camera movement actions.

To maximize the space occupation of $\mathbf{c}_\mathit{pred}$ in $\mathbf{v}_\mathit{pred}$, we can define the relationship between $\mathbf{v}_\mathit{pred}$ and $\mathbf{c}_\mathit{pred}$, \ie,
they share the same center position,
\begin{equation}
    (\mathit{x}^{\mathbf{v}}_\mathit{pred},\mathit{y}^{\mathbf{v}}_\mathit{pred})=(\mathit{x}^{\mathbf{c}}_\mathit{pred},\mathit{y}^{\mathbf{c}}_\mathit{pred}),
    \label{eqn:BIB_center_coordinate}
\end{equation}
and have the same width and/or height,
\begin{equation}
\begin{cases}
    \mathit{w}^{\mathbf{v}}_\mathit{pred}=\mathit{w}^{\mathbf{c}}_\mathit{pred}, & {\mathit{w}^{\mathbf{c}}_\mathit{pred}}/{\mathit{h}^{\mathbf{c}}_\mathit{pred}} \geq {\mathit{w}^{\mathbf{v}}_\mathit{pred}}/{\mathit{h}^{\mathbf{v}}_\mathit{pred}}\\
    \mathit{h}^{\mathbf{v}}_\mathit{pred}=\mathit{h}^{\mathbf{c}}_\mathit{pred}, & {\mathit{w}^{\mathbf{c}}_\mathit{pred}}/{\mathit{h}^{\mathbf{c}}_\mathit{pred}} \leq {\mathit{w}^{\mathbf{v}}_\mathit{pred}}/{\mathit{v}^{\mathbf{c}}_\mathit{pred}}
\end{cases},
\label{eqn:BIB_width_height}
\end{equation}
and $\mathbf{v}_\mathit{pred}$ will coincide with $\mathbf{c}_\mathit{pred}$ when they have the same aspect ratio.
Note that the camera view ratio is typically kept unchanged during the photography process, without loss of generality, in this paper we assume the camera view ratio to be $\mathit{w}^\mathbf{v}\!:\!\mathit{h}^\mathbf{v}=4\!:\!3$ or $\mathit{w}^\mathbf{v}\!:\!\mathit{h}^\mathbf{v}=3\!:\!4$.
Then given \cref{eqn:BIB_center_coordinate,eqn:BIB_width_height}, $\mathbf{v}_\mathit{pred}$ can be naturally derived from $\mathbf{c}_\mathit{pred}$.
Thus, we simplify the problem defined in \cref{eqn:BIB_definition} as,
\begin{equation}
    \mathbf{c}_\mathit{pred} = \mathit{f}(\mathbf{I}_\mathit{init}),
    \label{eqn:BIB_definition_simple}
\end{equation}
which can also be easily generalized to other camera view ratios or even adjustable ratios.

\subsection{Unbounded Regression Model}
\label{sec:base_model}
With the simplified task in \cref{eqn:BIB_definition_simple}, $\mathit{f}(\cdot)$ can be regarded as a generalized image cropping model which allows the predicted bounding box $\mathbf{c}_\mathit{pred}$ to exceed the image borders.
As such, we can implement the UNIC model based on existing image cropping models~\cite{guo2018automatic,lu2019end,li2018a2,li2019fast,hong2021composing}.
In particular, Jia~\etal~\cite{jia2022rethinking} have successfully applied DETR-like architectures~\cite{carion2020end,meng2021conditional} in image cropping tasks, which enables global interactions via the attention mechanism, and the set prediction settings also benefit the diversity of the results.
Therefore, we follow Jia~\etal~\cite{jia2022rethinking} and adopt conditional-DETR~\cite{meng2021conditional} as a base model for implementing $f(\cdot)$.

\vspace{0.5em}
\noindent\textbf{Network Design.}
In specific, as shown in \cref{fig:BIB_method}, following cDETR~\cite{meng2021conditional}, the base model contains a CNN backbone, a transformer encoder, a transformer decoder, as well as two heads $\mathit{H}_\mathit{pred}$ and $\mathit{H}_\mathit{conf}$ for predicting candidate bounding boxes and their corresponding confidence, respectively.
The initial view $\mathbf{I}_\mathit{init}$ is extracted into deep feature $\mathbf{h}_\mathit{init}$ with the CNN backbone, which is reorganized into patches.
Then the patches with positional embeddings attached according to their spatial positions are processed by the transformer encoder, resulting in a group of latent features denoted by $\mathcal{Z}_\mathit{vis}$.
Finally, the transformer decoder and two head branches predict candidate image composition results from a group of learnable anchors denoted by $\mathcal{A}=\{\mathbf{a}_1, \mathbf{a}_2,\dots,\mathbf{a}_\mathit{n}\}$.
Specifically, the bounding box head $\mathit{H}_\mathit{box}$ generates the coordinate of $\mathit{n}$ possible bounding boxes from the anchors (\ie, $\mathbf{c}_\mathit{pred}$), and the confidence head $\mathit{H}_\mathit{conf}$ predicts the confidence (or possibility) for each candidate bounding box (denoted by $\mathbf{p}_\mathit{pred}$).

However, as one can see, $\mathcal{Z}_\mathit{vis}$ only contains the feature of visible parts in the range of $\mathbf{v}_\mathit{init}$.
For improving the prediction accuracy beyond the initial camera view $\mathbf{v}_\mathit{init}$, a feature extrapolation module (FEM) is inserted into the base model.
The FEM is intended to predict the latent features outside $\mathbf{v}_\mathit{init}$, and the padded features are denoted by $\mathcal{Z}_\mathit{pad}$.
For predicting patches in $\mathcal{Z}_\mathit{pad}$, a learnable token $\mathbf{m}$ is fed into the FEM together with the positional embeddings.
We will give more details about the FEM in \cref{sec:extrapolation}.

\vspace{0.5em}
\noindent\textbf{Model Training.}
For training the UNIC model, we design the learning objective for composition mainly following Jia~\etal~\cite{jia2022rethinking}, \ie,
\begin{equation}
\begin{split}
    \mathcal{L}_\mathrm{comp} &= \mathcal{L}_\mathrm{reg}(\mathbf{c}_\mathit{pred}, \mathbf{c}) + \lambda_\mathrm{IoU}\mathcal{L}_\mathrm{IoU}(\mathbf{c}_\mathit{pred}, \mathbf{c}) \\
    &+ \lambda_\mathrm{focal}\mathcal{L}_\mathrm{focal}(\mathbf{p}_\mathit{pred}, \mathbf{p}),
\end{split}
\end{equation}
where $\lambda_\mathrm{IoU}$ and $\lambda_\mathrm{focal}$ are hyper-parameters for balancing different loss terms.
Note that there may exist multiple ground-truths for each $\mathbf{I}_\mathit{init}$, and the number of ground-truths might be different from the number of predicted bounding boxes.
Following \cite{jia2022rethinking,meng2021conditional,carion2020end}, we find the corresponding ground-truth for the predicted bounding boxes $\mathbf{c}_\mathit{pred}$ via bipartite matching.
In this way, only the results with a corresponding ground-truth contribute to the regression loss $\mathcal{L}_\mathrm{reg}$ and IoU loss $\mathcal{L}_\mathrm{IoU}$.

Another key factor is the construction of $\mathbf{p}$.
An intuitive way is to assign 1 or 0 according to the existence of ground-truth bounding box for the $\mathit{i}$-th prediction result.
Jia~\etal~\cite{jia2022rethinking} further proposed two strategies to generate soft labels for GAICD~\cite{zeng2019reliable} and CPC~\cite{wei2018good}, respectively.
In specific, the one for GAICD~\cite{zeng2019reliable} is a soft label according to the aesthetic score of the ground-truth (denoted by quality guidance), while for CPC~\cite{wei2018good} whose labels are more sparse, they use the prediction of the exponential moving averaged model as ground-truth (denoted by self-distillation).
In this work, we find that adopting the quality guidance strategy at first can stabilize the training process, and switching to the self-distillation strategy afterward further promotes the performance.
More details about the learning objectives are given in the supplementary material.

\subsection{Feature Extrapolation Module}
\label{sec:extrapolation}
To obtain an image composition result that may exceed the range of the initial view, Zhong~\etal~\cite{zhong2021aesthetic} expand the image by out-paining and predict the cropping scheme on the expanded image.
However, the out-painting manner may lead to unreal and inconsistent regions in the final image composition results, and there may be redundant or even harmful information in the generated pixels.
On the contrary, the space of the latent features $\mathcal{Z}_\mathit{vis}$ is dedicated to image composition tasks, which motivates us to extrapolate in the feature space of $\mathcal{Z}_\mathit{vis}$.

Recent advances in masked image modeling~\cite{chen2022context,zhang2022cae,he2022masked,xue2022stare,chen2022sdae} have achieved significant performance in predicting the representations of masked patches from visible parts of an image.
Inspired by their architectural design and learning schemes, we build our FEM module by stacking 6 transformer blocks.
The visible features $\mathcal{Z}_\mathit{vis}$ are involved in the feature extrapolation process through the cross-attention mechanism in the transformer block, and the detailed structure of the FEM is given in the supplementary material.

For training the FEM, it is insufficient if solely relies on the image composition loss $\mathcal{L}_\mathrm{comp}$.
In order to provide extra supervision for FEM, we leverage the full view image $\mathbf{I}$ (the initial view $\mathbf{I}_\mathit{init}$ is extracted from $\mathbf{I}$), and obtain the full view latent features $\mathcal{Z}$ via the CNN backbone and transformer encoder.
Then $\mathcal{Z}$ is split into two categories, \ie, $\mathcal{Z}_\mathit{in}$ in the range of $\mathbf{I}_\mathit{init}$ and $\mathcal{Z}_\mathit{out}$ outside $\mathbf{I}_\mathit{init}$.
As such, we can construct another supervision with $\mathcal{Z}_\mathit{out}$ for the extrapolation via FEM, where a robust smooth-$\ell_1$ loss~\cite{girshick2015fast} is adopted, \ie,
\begin{equation}
    \mathcal{L}_\mathrm{extra} = \mathrm{smooth}\textrm{-}{\ell_1}(\mathcal{Z}_\mathit{pad}, \mathit{sg}(\mathcal{Z}_\mathit{out})),
\end{equation}
where $\mathit{sg}(\cdot)$ is the stop gradient operator.
Note that to improve training stability, the parameters of the CNN backbone and transformer encoder for extracting $\mathcal{Z}$ are from the exponential moving averages (EMA) of the corresponding UNIC parameters.
The overall learnng objective for training the UNIC model is defined by,
\begin{equation}
    \mathcal{L}=\mathcal{L}_\mathrm{comp} + \mathcal{L}_\mathrm{extra}.
\end{equation}

\subsection{Unbounded Image Composition Dataset}
\label{sec:dataset}
Even though there are several datasets~\cite{wei2018good,zeng2019reliable,chen2017quantitative} for image composition tasks, all of them are intended for cropping based image composition tasks, and there is no publicly available dataset for unbounded image composition.
To make full use of the aesthetic annotations in existing image cropping datasets, we recreate an unbounded image composition dataset based on GAICD~\cite{zeng2019reliable} and CPC~\cite{wei2018good}.

In specific, for a sample in image cropping datasets, a full-view image is provided with one or multiple ground-truth bounding boxes.
All these ground-truths are located in the range of the full-view image, making it infeasible for unbounded image composition tasks.
As a remedy, we randomly sample a bounding box (\ie, $\mathbf{v}_\mathit{init}$) in the full-view image, then the ground-truths may not fully lie in the range of $\mathbf{v}_\mathit{init}$.
In other words, the ground-truths for cropping based image composition are adapted to unbounded image composition tasks.

However, randomly sampling $\mathbf{v}_\mathit{init}$ with no constraints may be improper in particular situations.
For example, if the interested object is outside of $\mathbf{v}_\mathit{init}$, it is unreasonable to require that the predicted bounding box can cover that object.
Therefore, we set up some rules as follows when recreating the unbounded image composition dataset.

\vspace{0.5em}
\noindent\textbf{Size of $\mathbf{I}_\mathit{init}$}.
To ensure the initial camera view contains sufficient image content, we set the lower bound of the height and width of $\mathbf{I}_\mathit{init}$ as
\begin{equation}
\mathit{h}^\mathbf{v}_\mathit{init}\geq\alpha\cdot\mathit{h} \  \textrm{and}\ \mathit{w}^\mathbf{v}_\mathit{init}\geq\alpha\cdot\mathit{w},
\end{equation}
where $\mathit{h}$ and $\mathit{w}$ denote the height and width of full-view image $\mathbf{I}$, and $\alpha$ is the scale threshold empirically set to 0.7.

\vspace{0.5em}
\noindent\textbf{Position of $\mathbf{I}_\mathit{init}$}.
Apart from high aesthetic qualities, an important property of the ground-truth bounding boxes is that they well describe the range of desired objects or scenes.
To ensure that the initial view contains the desired objects or scenes, we constrain the intersection of union (IoU) of the initial view $\mathbf{v}_\mathit{init}$ and the ground-truth $\mathbf{v}$.
Specifically, the constraint is defined as,
\begin{equation}
\mathrm{IoU}(\mathbf{v}_\mathit{init}, \mathbf{v}) \geq \beta,
\end{equation}
where the threshold $\beta$ is set to 0.7 in this paper.

\vspace{0.5em}
\noindent\textbf{Aspect ratio of $\mathbf{I}_\mathit{init}$}.
Considering that the camera view ratio is typically kept unchanged during the photography process, without loss of generality, we sample $\mathbf{I}_\mathit{init}$ with an aspect ratio of $4\!:\!3$, which is the most common setting for DLSRs and smartphones.

Since the cameras could take photos vertically or horizontally, we have
\begin{equation}
    \mathit{w}^\mathbf{v}_\mathit{init}\!:\!\mathit{h}^\mathbf{v}_\mathit{init}=4\!:\!3\ \ \textrm{or}\ \ \mathit{w}^\mathbf{v}_\mathit{init}\!:\!\mathit{h}^\mathbf{v}_\mathit{init}=3\!:\!4.
\end{equation}

\section{Experiments}

\begin{table*}
\centering
\caption{Quantitative comparison for unbounded image composition on GAICD~\cite{zeng2019reliable} and FLMS~\cite{fang2014automatic} datasets. The best results are highlighted with \textbf{bold}. The results marked by $\dagger$ and $\ddagger$ are retrained with our data or reproduced in our framework, respectively.}
\vspace{1mm}
\begin{tabular}{p{2.7cm}<{\centering} | p{1.3cm}<{\centering} p{1.3cm}<{\centering} p{1.3cm}<{\centering} p{1.3cm}<{\centering}  p{1.1cm}<{\centering}  p{1.1cm}<{\centering} | p{1.1cm}<{\centering}  p{1.1cm}<{\centering}}
\hline
\multirow{3}{*}{Method} &\multicolumn{6}{c|}{GAICD} &\multicolumn{2}{c}{FLMS} \\
\cline{2-9}

& \multicolumn{2}{c|}{$Acc_{1/5}$} & \multicolumn{2}{c|}{$Acc_{1/10}$} & \multicolumn{1}{c}{\multirow{2}{*}{IoU  $\uparrow$}} & \multicolumn{1}{c|}{\multirow{2}{*}{Disp  $\downarrow$ }} & \multirow{2}{*}{IoU  $\uparrow$} & \multirow{2}{*}{Disp  $\downarrow$ } \\ 
\cline{2-3}
\cline{4-5}
& $\epsilon=$0.90 & \multicolumn{1}{c|}{$\epsilon=$0.85}  & $\epsilon=$0.90 &  \multicolumn{1}{c|}{$\epsilon=$0.85} & &\multicolumn{1}{c|}{} &  &  \\ 
\hline
VFN~\cite{chen2017learning} & 0.6 & 5.2 & 1.7 & 9.5 & 0.577 & 0.124 & 0.622 & 0.122\\
VEN~\cite{wei2018good}  & 2.6 & 8.9 & 3.4 & 11.5 & 0.600 & 0.095 &0.688 & 0.065\\
GAIC~\cite{zeng2019reliable} & 7.2 & 21.8 & 10.6 & 31.5 & 0.683 & 0.074 & 0.723 &0.060
\\
CGS~\cite{li2020composing} & 7.2 & 25.8 & 10.9 & 33.5 & 0.682 & 0.074  & 0.703 & 0.064\\
\hline
A2-RL~\cite{li2018a2} & 6.9 & 22.9 & 11.2 & 34.7 & 0.686 & 0.076 & 0.731& 0.059 \\
CACNet$^{\dagger}$~\cite{hong2021composing} & 16.9 & 49.1 & 25.8 & 60.7 & 0.779 & 0.052 & 0.813 & 0.044
\\
Zhong~\etal$^\ddagger$~\cite{zhong2021aesthetic} & 22.3 & 53.5 & 28.7 & 67.2 & 0.795 & 0.050 & 0.818 & 0.042
\\
Jia~\etal$^\dagger$~\cite{jia2022rethinking} & 21.4 & 48.0 & 26.8 & 57.2 & 0.786 & 0.052 & 0.817 & 0.042\\
\hline
Ours & \textbf{23.2} & \textbf{59.0} & \textbf{32.7} & \textbf{72.8} & \textbf{0.801} & \textbf{0.048} & \textbf{0.828} & \textbf{0.040}\\
\hline
\end{tabular}
\label{tab:comparisonview}
\end{table*}

\subsection{Implementation Details}
\noindent\textbf{Datasets}.
We adopt two widely-used datasets for training, \ie, GAICD~\cite{zeng2019reliable} and CPC~\cite{wei2018good}.
GAICD~\cite{zeng2019reliable} is a grid anchor based image cropping dataset, where each image has exhaustive annotations for the cropping candidates.
It contains 2,636 images for training, 200 images for validation, and 500 images for testing.
We train our model on the training split and evaluate it on the testing split.
CPC~\cite{wei2018good} dataset is sparsely annotated for training purposes only, which contains 10,800 images with 24 annotated views per image.
We evaluate our model trained with CPC~\cite{wei2018good} on FLMS dataset~\cite{fang2014automatic} following \cite{jia2022rethinking}.
Both datasets are pre-processed for unbounded image composition as shown in \cref{sec:dataset}.

\vspace{0.5em}
\noindent\textbf{Evaluation metrics}.
The camera view recommendation accuracy could be measured with the intersection of union (IoU) and boundary displacement (Disp) between the predicted view and the ground-truth view with the highest aesthetic score following~\cite{wei2018good}.
However, there may exist multiple human-annotated ground-truth bounding boxes in each image, these metrics ignore such situations, which limits their flexibility.
As a remedy, $\mathit{Acc}_{\mathit{K}\!/\!\mathit{N}}$ calculates how many of $\mathit{K}$ predicted views falls into the $\mathit{N}$ ground-truth views with highest score.
Therefore, we adopt $\mathit{Acc}_{\mathit{K}\!/\!\mathit{N}}$ as another evaluation metric for grid annotated GAICD~\cite{zeng2019reliable}.
As the predicted views may not align perfectly with the pre-defined grid views, we follow~\cite{jia2022rethinking} and regard two crops the same when their IoU is sufficiently large.
Two thresholds $\epsilon=\{0.9, 0.85\}$ are used in this paper.
For FLMS dataset~\cite{fang2014automatic} without grid annotation, we use IoU and Disp metrics.

\vspace{0.5em}
\noindent\textbf{Training details}.
The amount of data in the image cropping datasets is not sufficiently enough for training DETR-like models from scratch.
Thus, we initialize the CNN backbone with ImageNet pre-trained weights~\cite{he2016deep}.
The layer numbers of the transformer encoder and decoder are both set to 6.
During training, we take views with an aesthetic score larger than 4 in GAICD~\cite{zeng2019reliable} and that larger than 2 in CPC dataset~\cite{wei2018good} as ground-truth views.
The trade-off parameters $\lambda_\mathrm{IoU}$ and $\lambda_\mathrm{focal}$ are set to 0.4 and 0.1, respectively.
The model is trained with an AdamW~\cite{loshchilov2017decoupled} optimizer with weight decay of $1\times10^{-4}$ for 50 epochs.
The initial learning rates for the CNN backbone and the transformer encoder/decoder are set to $1\times10^{-5}$ and $1\times10^{-4}$, which are decreased to 1/10 at the 30-th epoch. 
We apply data augmentation via color jittering and resizing following~\cite{meng2021conditional}.
%

\subsection{Results of Unbounded Image Composition}
Due to the lack of competing methods for unbounded image composition, we adopt several state-of-the-art image cropping methods with source code publicly available, including anchor evaluation based methods, \ie, VFN~\cite{chen2017learning}, VEN~\cite{wei2018good}, GAIC~\cite{zeng2019reliable}, and CGS~\cite{li2020composing}, as well as coordinate regression based methods, \ie, A2-RL~\cite{li2018a2}, CACNet~\cite{hong2021composing}, and Jia~\textit{et~al.}~\cite{jia2022rethinking}.
Among them, the anchor evaluation based methods require the cropped image for scoring.
Directly applying them for unbounded image composition tasks will lead to poor results due to the incomplete image for views exceeding the initial view borders.
Therefore, we show the results of cropping based image composition for these anchor evaluation based methods.
As for coordinate regression based methods, we show cropping based results of A2-RL~\cite{li2018a2} since it is based on VFN~\cite{chen2017learning}, and retrain CACNet~\cite{hong2021composing} and the method of Jia~\etal~\cite{jia2022rethinking} with our training data.
Zhong~\etal~\cite{zhong2021aesthetic} can predict cropping schemes via image extrapolation, which is the most similar method to our UNIC.
Since the source code is unavailable, we reimplement their method in our framework, where the extrapolation module is replaced by a StyleGAN2~\cite{karras2020analyzing} based image out-painting model~\cite{zhao2021large}.

\vspace{0.5em}
\noindent\textbf{Quantitative comparison.}
We conduct comprehensive experiments to assess the effectiveness of the proposed method, and the quantitative results are shown in \cref{tab:comparisonview}.
From the $\mathit{Acc}_{1/5}$ and $\mathit{Acc}_{1/10}$ metrics with two IoU thresholds  $\epsilon\in\{0.90, 0.85\}$ on GAICD~\cite{zeng2019reliable}, we can see that anchor evaluation based methods~\cite{chen2017learning, wei2018good, zeng2019reliable, li2020composing} are limited by the border of the current view.
Regression based methods~\cite{li2018a2, hong2021composing, jia2022rethinking} show inferior results as they are not properly designed for unbounded image composition tasks.
Our method instead shows significant improvement for unbounded image composition tasks compared to the previous methods.
The IoU and Disp metrics in GAICD ~\cite{zeng2019reliable} and FLMS ~\cite{fang2014automatic} datasets also demonstrate the effectiveness of the proposed UNIC model.

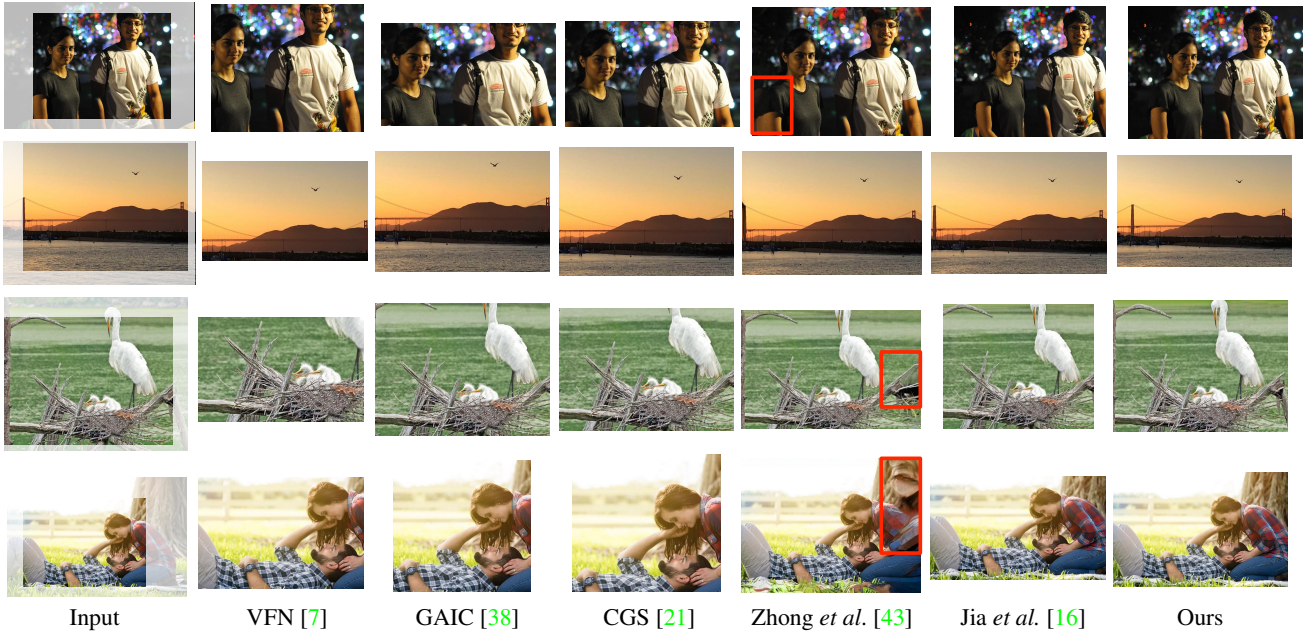
\begin{figure*}[htp]
\centering
 \begin{overpic}[width=0.99\textwidth]{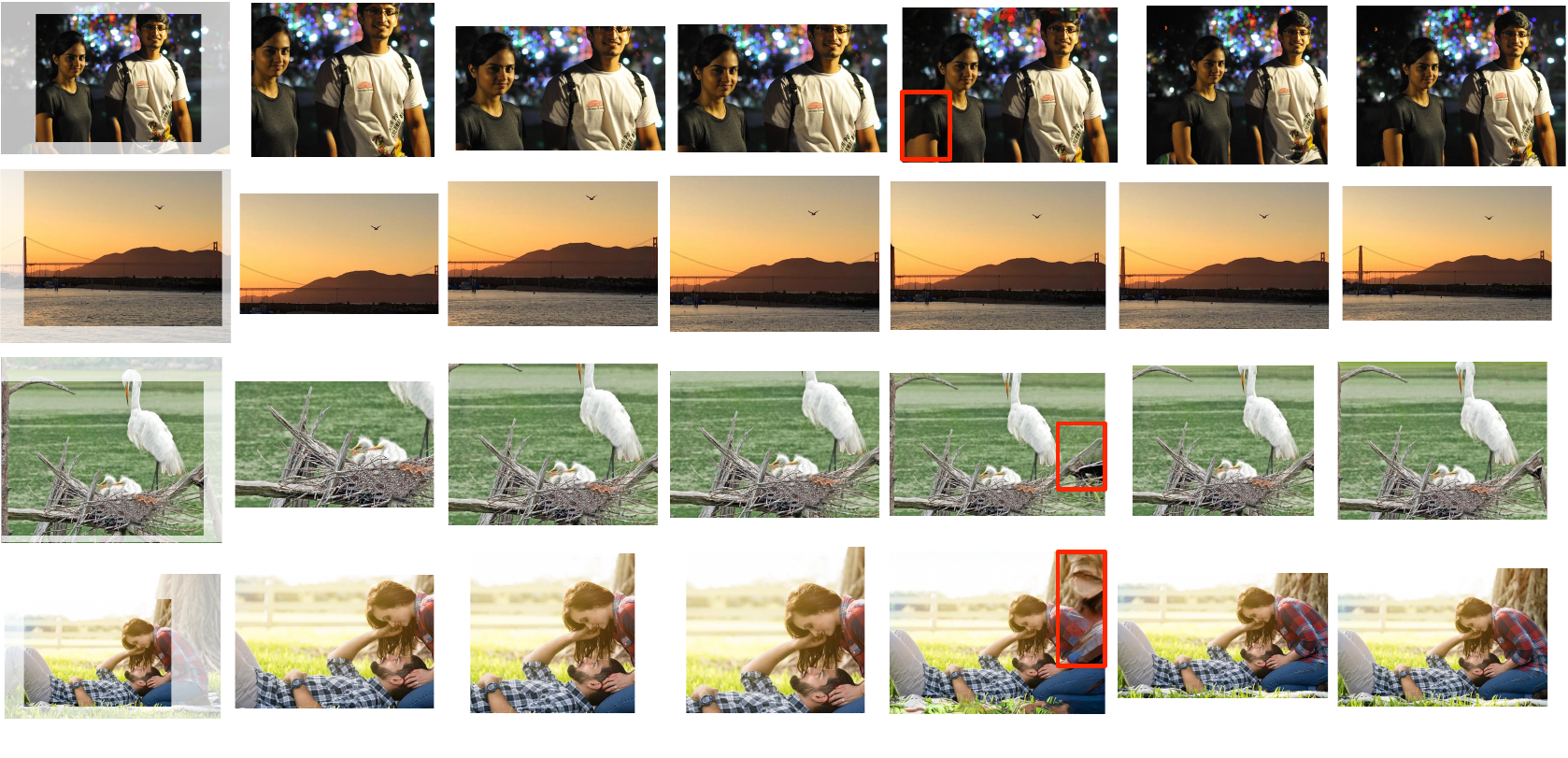}
  \put(25,1.5){\small{Input}}
  \put(92,1.5){\small{VFN~\cite{chen2017learning}}}
  \put(156,1.5){\small{GAIC~\cite{zeng2019reliable}}}
  \put(227,1.5){\small{CGS~\cite{li2020composing}}}
  \put(283,1.5){\small{Zhong~\etal~\cite{zhong2021aesthetic}}}
  \put(362,1.6){\small{Jia~\textit{et~al.}~\cite{jia2022rethinking}}}
  \put(444,1.5){\small{Ours}}
\end{overpic}
\vspace{1mm}
\caption{Qualitative comparison with other methods.
Our method goes beyond the border of the image to predict a well-composed region with the main objects in reasonable places.}
\label{fig:BIB_compare}
\end{figure*}

\vspace{0.5em}
\noindent\textbf{Qualitative comparison}.
The qualitative results of competing methods are shown in \cref{fig:BIB_compare}.
%
%
Anchor evaluation methods~\cite{chen2017learning,zeng2019reliable, li2020composing} are restricted by the boundary of the initial view, which cannot adjust the camera toward a larger view and show inferior results.
After training with our dataset, the regression based methods~\cite{jia2022rethinking} could predict outward views, but due to the implicit regression from inbound contents, their accuracy in the out-of-border regions is also limited.
For \cite{zhong2021aesthetic}, the extrapolated regions may be included in the cropped image, which harms the aesthetic quality such as the abnormal arm in the first row and the artifacts near the woman in the fourth row.
In contrast, our method not only learns to predict beyond image borders, but also predicts a more accurate and aesthetically pleasurable view by extrapolation in the feature space.
More qualitative results are given in the supplementary material.

\newcommand{\tabincell}[2]{\begin{tabular}{@{}#1@{}}#2\end{tabular}}

\begin{table}
\centering
\caption{Quantitative comparison for image cropping on GAICD~\cite{zeng2019reliable} and FLMS~\cite{fang2014automatic} datasets. $\mathit{Acc}_{1/5}$ and $\mathit{Acc}_{1/10}$ are calculated according to~\cite{jia2022rethinking}.}
\vspace{1mm}
\begin{tabular}{ p{3.0cm}<{\centering} | p{0.82cm}<{\centering} p{0.92cm}<{\centering} | p{0.81cm}<{\centering} p{0.92cm}<{\centering}}
\hline
  \multirow{2}{*}{Method} &\multicolumn{2}{c|}{GAICD} &\multicolumn{2}{c}{FLMS} \\
\cline{2-5}
   & $\mathit{Acc}_{1/5}$ & $\mathit{Acc}_{1/10}$ & IoU $\uparrow$  & Disp $\downarrow$  \\
\hline
  VFN ~\cite{chen2017learning}  & 26.6 & 40.6  & 0.577 & 0.124\\
  VEN ~\cite{wei2018good}  & 37.5 & 48.5 & 0.837 & 0.041\\
  GAIC ~\cite{zeng2019reliable} & 68.2 & 85.8  & 0.834 & 0.041\\
  CGS ~\cite{li2020composing} & 63.0 & 81.5  & 0.836 & 0.039 \\
\hline
  A2-RL~\cite{li2018a2}~($\epsilon$=0.90)  & 7.6 & 12.6 & \multirow{2}{*}{0.821} & \multirow{2}{*}{0.045}\\
  A2-RL~\cite{li2018a2}~($\epsilon$=0.85)  & 28.6 & 43.2 \\
  \hline
CACNet~\cite{hong2021composing}~($\epsilon$=0.90)  & 50.7 & 66.0  & \multirow{2}{*}{\textbf{0.854}} & \multirow{2}{*}{\textbf{0.033}}\\
CACNet~\cite{hong2021composing}~($\epsilon$=0.85)  & 78.0 & 89.3 \\
\hline
 {Jia~\textit{et~al.}~\cite{jia2022rethinking}~($\epsilon$=0.90)} & 72.0 & 86.0  & \multirow{2}{*}{0.838} & \multirow{2}{*}{0.037}\\
 {Jia~\textit{et~al.}~\cite{jia2022rethinking}~($\epsilon$=0.85)} & 85.0  & 92.6  \\
\hline
{Ours~($\epsilon$=0.90)}  & \textbf{74.7} & \textbf{89.6}  & \multirow{2}{*}{0.840} & \multirow{2}{*}{0.037}\\
{Ours~($\epsilon$=0.85)}  &\textbf{87.2} & \textbf{95.5}  \\
\hline
\end{tabular}
\label{tab:comparisonframe}
\end{table}

\subsection{Results for Image Cropping}
Although our UNIC is delicately designed for unbounded image composition tasks, it can degrade to an image cropping model with the absence of the FEM.
%
%
\cref{tab:comparisonframe} shows the results for image cropping task on the original GAICD~\cite{zeng2019reliable} and FLMS~\cite{fang2014automatic} datasets.
One can see that our method outperforms all existing methods that are specifically designed for image cropping tasks on the GAICD~\cite{zeng2019reliable} dataset and achieves comparable performance to the state-of-the-art methods on the FLMS~\cite{fang2014automatic} dataset, showing the effectiveness of the proposed UNIC framework.

\begin{table}[t]
\centering
\caption{Ablation study on the extrapolation (Extra.) strategies. Extrapolation in the feature space achieves the best results.}
\vspace{1mm}
\begin{tabular}{p{4.2cm}<{\centering} | p{1.4cm}<{\centering} | p{1.4cm}<{\centering} }
\hline
Method & \tabincell{c}{$\mathit{Acc}_{1/5}$ \\ $(\epsilon=0.90)$} & \tabincell{c}{$\mathit{Acc}_{1/5}$ \\ $(\epsilon=0.85)$} \\
\hline
Ours w/o Extra. & 22.6 & 48.1 \\
\hline
Ours w/ SRN~\cite{wang2019wide} & 19.9 & 51.0 \\
Zhong~\etal~\cite{zhong2021aesthetic} & 22.3 & 53.5 \\
Ours w/ QueryOTR~\cite{yao2022outpainting} & 23.1  & 53.7 \\
\hline
Ours w/ Feature Extra. & \textbf{23.2} & \textbf{59.0} \\
\hline
\end{tabular}
\label{tab:extrapolation}
\end{table}

\subsection{Ablation Study}
\noindent\textbf{Effects of extrapolation strategy}.
As illustrated in \cref{sec:extrapolation}, we apply extrapolation in the feature space to boost the performance of our unbounded regression model.
In this subsection, we make detailed experiments to assess the effects of different extrapolation strategies, \eg, no extrapolation, image extrapolation, and feature extrapolation.
We take the UNIC without FEM as the model with no extrapolation, and several state-of-the-art out-painting methods~\cite{wang2019wide, zhao2021large, yao2022outpainting} are applied to the input image for evaluating the image-level extrapolation.
As shown in \cref{tab:extrapolation}, image-level extrapolation may suffer from generative artifacts due to the extrapolation model, as our model with SRN~\cite{wang2019wide} exhibits a performance drop in $\mathit{Acc}_{1/5}\ (\epsilon=0.90)$ and limited improvement on $\mathit{Acc}_{1/5}\ (\epsilon=0.85)$.
With more powerful generative models~\cite{karras2020analyzing, yao2022outpainting}, image-level extrapolation shows consistent improvement.
Nonetheless, our model with feature extrapolation benefits from recent advances in mask image modeling~\cite{chen2022context} and end-to-end training, which shows the best results.
It achieves a 2.5\% improvement on $\mathit{Acc}_{1/5}\ (\epsilon=0.90)$ and a 22.6\% improvement on $\mathit{Acc}_{1/5}\ (\epsilon=0.85)$ over the base model, which demonstrates the effectiveness of extrapolation in the feature space.
More analysis and visual results are provided in the supplementary material.

\vspace{0.5em}
\noindent\textbf{Effects of FEM loss}.
In order to assess the effects on the loss function of the FEM for feature extrapolation, we experiment on several commonly used loss functions for regression, \eg, mean square error (MSE), cosine distance, KL-divergence, and smooth-$\ell_1$.
As shown in \cref{tab:loss}, smooth-$\ell_1$ yields the best overall performance.
Thus we choose smooth-$\ell_1$ loss for our FEM in this paper.

\begin{table}
\centering
\caption{Ablation study on the $\mathcal{L}_\mathrm{extra}$ for feature extrapolation.}
\vspace{1mm}
\begin{tabular}{p{3.4cm}<{\centering} | p{1.6cm}<{\centering} | p{1.6cm}<{\centering} }
\hline
Type & \tabincell{c}{$\mathit{Acc}_{1/5}$ \\ $(\epsilon=0.90)$} & \tabincell{c}{$\mathit{Acc}_{1/5}$ \\ $(\epsilon=0.85)$} \\
\hline
MSE & 22.1 & 56.1 \\
Cosine Distance & 23.1 & 57.9 \\
KL-Divergence & \textbf{23.8} & 56.4 \\
Smooth-$\mathcal{L}_1$ & 23.2  & \textbf{59.0} \\
\hline
\end{tabular}
\label{tab:loss}
\end{table}

\begin{table}
\centering
\caption{Ablation study on multi-step adjustment.}
\vspace{1mm}
\begin{tabular}{c| c | c| c}
\hline
 & step=1 & step=2 & step=3 \\
\hline
 $\mathit{Acc}_{1/5}\ (\epsilon=0.90)$ & 16.9 & 19.3 & \textbf{19.3} \\
  $\mathit{Acc}_{1/5}\ (\epsilon=0.85)$ & 48.2 & 51.8 & \textbf{54.2} \\
\hline
\end{tabular}
\label{tab:step}
\end{table}

\begin{figure}[t]

	\begin{minipage}{0.24\linewidth}
        \centering
		\centering{\small{Input}}
        \includegraphics[width=\textwidth]{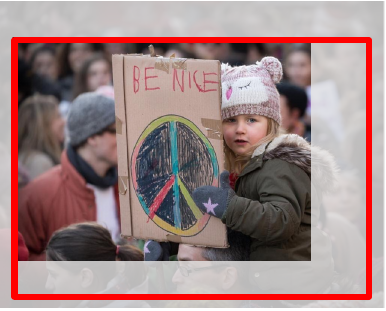}
		\centering{\small{IoU=0.721}}
    \end{minipage}   
    \begin{minipage}{0.24\linewidth}
        \centering
        \centering{\small{Step=1}}
    	\includegraphics[width=\textwidth]{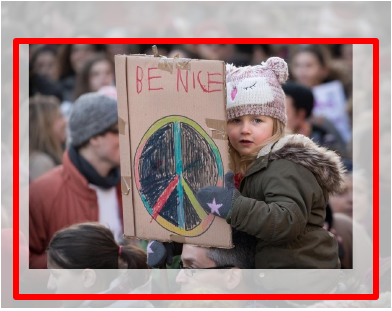}
		\centering{\small{IoU=0.810}}
    \end{minipage}
    \begin{minipage}{0.24\linewidth}
        \centering
        \centering{\small{Step=2}}
		\includegraphics[width=\textwidth]{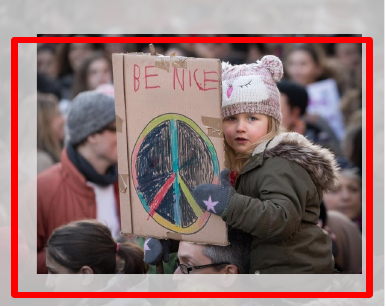}
		\centering{\small{IoU=0.844}}
    \end{minipage}
    \begin{minipage}{0.24\linewidth}
        \centering
        \centering{\small{Step=3}}
		\includegraphics[width=\textwidth]{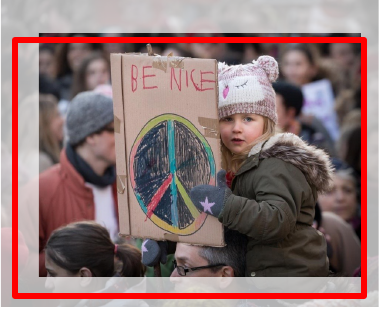}
		\centering{\small{IoU=0.852}}
	\end{minipage}
 
    \vspace{2mm}
\caption{Visual comparison for multi-step adjustment. Our model predicts better view with increased adjust steps to approach the ground-truth view within the red box.}
\label{fig:BIB_step_fig}
\end{figure}

\vspace{0.5em}
\noindent\textbf{Effects of multi-step adjustment}.
The camera view predicted from our regression model may not be the most aesthetic view with the unseen regions, but is expected to move closer toward it.
Based on the above idea, the camera view could be further improved with multi-step adjustment.
Concretely, we first apply our model to the initial view captured by the camera, predict camera operations, and perform adjustments.
Then the same process is applied on the new view after adjustment, which could be operated multiple times.
\cref{tab:step} shows the results of multi-step adjustment.
We note that we use images in GAICD~\cite{zeng2019reliable} as the whole scene and a crop within the images as the initial view in our experiment.
Multi-step adjustment may exceed the border of the scene, so we select 83 large images from GAICD~\cite{zeng2019reliable} to avoid this problem, thus the results in \cref{tab:step} are not consistent with other tables in the paper.
From the table, the performance is promoted with increased adjustment steps, which demonstrates the effectiveness of the multi-step adjustment.
As shown in \cref{fig:BIB_step_fig}, our model predicts better views with increased adjustment steps to approach the ground-truth view circled by the red box.

\begin{figure}[t]
\centering
    \begin{overpic}[width=0.49\textwidth]{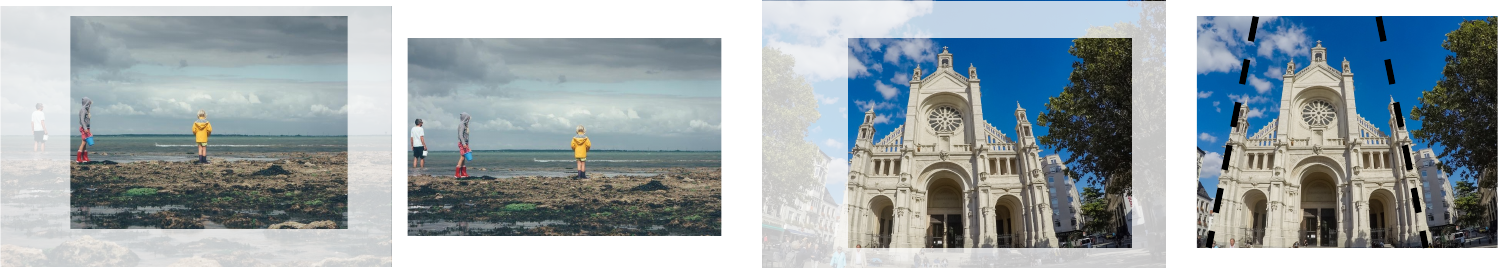}
    \end{overpic}
\vspace{-3mm}
\caption{Visualization of failure cases.}
\label{fig:BIB_fail}
\end{figure}

\section{Limitation and Future Work}
Although UNIC predicts well-composed views in most scenarios, it may encounter failure in certain circumstances.
As shown in the left of \cref{fig:BIB_fail}, an unexpected people near
the border appears in the predicted view, which is unseen in the initial camera view and may the affect the aesthetics quality of the predicted view.
This could be addressed with multi-step adjustment as the predicted view becomes stable.
The right example shows the camera view adjustment operations are limited to shifting and zooming in or out in this paper, it's hard to adjust the camera view without camera pose adjustment.
Besides, more scene information (\eg, depth) could be leveraged for better camera view recommendation.
We leave these problems as future work.

\section{Conclusion}
In this paper, we propose a novel framework for \textbf{UN}bounded \textbf{I}mage \textbf{C}omposition, \ie, UNIC.
Different from previous image cropping methods that improve the composition in a post-process manner, UNIC provides recommendations for camera view adjustment during photographing.
To improve the model accuracy beyond borders, we introduce a feature extrapolation module based on recent advances in mask image modeling.
To assist the model training and evaluation, we construct unbounded image composition datasets based on existing image cropping ones.
Extensive experiments demonstrate that our UNIC achieves better performance against the state-of-the-art methods in both image cropping and unbounded image composition tasks.

\section*{Acknowledgement}
This work was supported in part by the National Key Research and Development Program of China under Grant No. 2022YFA1004103 and the National Natural Science Foundation of China (NSFC) under Grant No. U19A2073.

{\small
\bibliographystyle{ieee_fullname}
\bibliography{egpaper}
}

\end{document}